\documentclass{bmvc2k}

\usepackage{amsmath,amssymb}
\usepackage{mathtools}

\usepackage{mathtools}
\usepackage{booktabs}
\usepackage{epsfig}
\usepackage{csquotes}
\usepackage{hyperref}

\title{Adverse Weather Image Translation with Asymmetric and Uncertainty-aware GAN}

\addauthor{Jeong-gi Kwak}{kjk8557@korea.ac.kr}{1}
\addauthor{Youngsaeng Jin}{youngsjin@korea.ac.kr}{1}
\addauthor{Yuanming Li}{lym7499500@korea.ac.kr}{1}
\addauthor{Dongsik Yoon}{kevinds1106@korea.ac.kr}{1}
\addauthor{Donghyeon Kim}{kis6470@korea.ac.kr}{1}
\addauthor{Hanseok Ko}{hsko@korea.ac.kr}{1}

\addinstitution{
 Korea Univerity\\
 Seoul, Korea
}

\runninghead{J.Kwak \etal}{AU-GAN: Asymmetric and Uncertainty-aware GAN}

\def\eg{\emph{e.g}\bmvaOneDot}

\def\etal{\emph{et al}\bmvaOneDot}

\begin{document}

\maketitle

\begin{abstract}
Adverse weather image translation belongs to the unsupervised image-to-image (I2I) translation task which aims to transfer adverse condition domain (\eg, rainy night) to standard domain (\eg, day). It is a challenging task because images from adverse domains have some artifacts and insufficient information. Recently, many studies employing Generative Adversarial Networks (GANs) have achieved notable success in I2I translation but there are still limitations in applying them to adverse weather enhancement. Symmetric architecture based on bidirectional cycle-consistency loss is adopted as a standard framework for unsupervised domain transfer methods. However, it can lead to inferior translation result if the two domains have {\em imbalanced information}. To address this issue, we propose a novel GAN model, i.e., \textit{AU-GAN}, which has an asymmetric architecture for adverse domain translation. We insert a proposed feature transfer network (${T}$-net) in only a normal domain generator (i.e., rainy night $\rightarrow$ day) to enhance encoded features of the adverse domain image. In addition, we introduce asymmetric feature matching for disentanglement of encoded features. Finally, we propose uncertainty-aware cycle-consistency loss to address the regional uncertainty of a cyclic reconstructed image. We demonstrate the effectiveness of our method by qualitative and quantitative comparisons with state-of-the-art models. Codes are available at \url{https://github.com/jgkwak95/AU-GAN}.      
\end{abstract}

\section{Introduction}
\label{sec:intro}

Thanks to the remarkable representation power and optimization technique of recent deep learning algorithms, there have been notable achievements in scene understanding tasks such as semantic segmentation~\cite{zhao2017pyramid, chen2017deeplab, chen2018encoder, jin2021trseg} and object detection~\cite{ren2015faster,tan2020efficientdet, bochkovskiy2020yolov4}. 
However, a number of such algorithms suffer from performance drop in a real world application. 
For example, semantic segmentation models trained on a dataset consisting mostly of day images show inferior results on night images.
The situation becomes even worse in adverse weather conditions such as rainy nights.
It could be the seeds of disaster in several applications \eg, autonomous driving in which the reliability of the algorithm is a critical factor. For this reason, there have been several approaches~\cite{jiang2021enlightengan, zheng2020forkgan} trying to transfer a challenging domain to a specific domain, where off-the-shelf methods work well. 

Since the emergence of Generative Adversarial Networks (GANs) by Goodfellow \etal{}~\cite{goodfellow2014generative}, conditional GAN (cGAN)~\cite{mirza2014conditional} has proposed its potential to be used in various generative tasks. 
Some variants of cGAN~\cite{Isola_2017_CVPR, wang2018high} have exploited an image as condition, i.e., image-to-image (I2I) translation.    
However these early I2I translation methods are not suitable for unsupervised domain transfer because they require ground truth pair for each image that is notoriously challenging to obtain.    
CycleGAN~\cite{Zhu_2017} resolves the unsupervised domain transfer problem by utilizing cycle-consistency loss and presents excellent translation results between unsupervised two domains. 
Because of its guaranteed performance, even the latest studies on unsupervised image-to-image translation include the cycle-consistency loss in their objective functions.
The crucial ability for domain translation model is to alter only domain specific factors (\eg, style or texture) while preserving domain invariant factors (\eg, object).     
To this end, several approaches~\cite{huang2018multimodal, lee2018diverse, liu2017unsupervised} have been proposed to disentangle domain-invariant and domain-specific features from two different domains by adopting concept of shared content feature space from two different domains. 
Lately, Zheng \etal{}~\cite{zheng2020forkgan} have proposed ForkGAN, that consists of cyclic translation with a "common encoding space" to disentangle domain invariant information. To this end, they adopt perceptual loss~\cite{simonyan2014very} between encoded features from two different domain. ForkGAN has shown reasonable translation results from adverse domain (rainy night) to normal domain (day). 

However, the symmetric architecture that commonly used in CycleGAN-based methods including ForkGAN would be inappropriate for adverse domain translation.
This is because there is a noticeable domain gap between standard and adverse weather. In other words, there are a lot of artifacts, blur, and reflections in rainy night images. 

To address the issues, we first introduce an asymmetric architecture for adverse domain translation. Here, only a generator that transfers adverse domain to standard domain has an additional network, i.e., feature transfer network, between an encoder and a decoder. The transfer network plays a role of enhancing the feature encoded from adverse domain images. We also introduce asymmetric feature matching loss to achieve better disentanglement without removing local objects. Li~\etal{}~\cite{li2019asymmetric} have approached in a similar concept in terms of "asymmetric", but their method does not consider the shared space of disentangled features from different domains for unsupervised image translation.  

Although the cycle-consistency loss helps to preserve the shape of the original image because of its powerful constraint in general, artifacts could remain in the case of the adverse domain. Motivated by uncertainty modeling~\cite{kendall2017uncertainties}, we introduce an uncertainty-aware cycle-consistency loss to alleviate the side-effect of the cycle-consistency loss. Through modeling uncertainty, the modified cyclic loss penalizes the regions of an image differently according to the confidence map. We analyze the effectiveness of our model with qualitative and quantitative experiments. 

Therefore, our contributions are as follow, 
\begin{itemize}
    \item We present a novel asymmetric GAN framework for adverse domain translation by utilizing a feature transfer network for one-way translation. 
    \item We introduce asymmetric feature matching loss and uncertainty-aware cycle consistency loss designed to consider the characteristics of images in the adverse domain.
    \item We demonstrate the superiority of our model by qualitative and quantitative comparisons with the state-of-the-art methods.
\end{itemize}

\section{Related work}
\paragraph{Unsupervised image-to-image translation} Since the introduction of GAN by Goodfellow \etal~\cite{goodfellow2014generative}, there have been numerous studies on Image-to-image translation (I2I) which aims to transfer an image of the source domain to that of the desired target domain. 
CycleGAN~\cite{Zhu_2017} proposed cycle-consistency loss for translation between unsupervised source and target domain. 
This concept has influenced several unsupervised domain translation tasks such as face attribute editing~\cite{He2017AttGANFA, Choi_2018, Liu_2019_CVPR, kwak2020cafe} or domain adaptation~\cite{murez2018image,Tzeng_2017_CVPR, liu2018unified}. StarGAN~\cite{Choi_2018} and AttGAN~\cite{He2017AttGANFA} conduct multi-domain translation by adopting a target vector of desired attributes as an additional input. 
UNIT~\cite{liu2017unsupervised} brings the concept of the shared latent space of two generators via weight sharing.   
To produce diverse outputs, MUNIT~\cite{huang2018multimodal} and DRIT~\cite{lee2018diverse} suggest and develop the concept of disentangled representation by decomposing an image into two spaces, i.e., shared domain-invariant space and domain specific space. Furthermore, several recent studies~\cite{lee2020drit++, choi2020stargan} present multi-modal outputs in multi-domain by exploiting disentanglement assumption. 
         
\paragraph{Adverse weather enhancement} 
Numerous models related to scene understanding vision tasks such as semantic segmentation and object detection suffer from degraded performance in bad weather conditions. This is because they are trained with a dataset composed mostly of normal weather images (\eg, daytime). As generative models evolve, there have been several attempts to enhance adverse weather images by the I2I translation technique. EnlightenGAN~\cite{jiang2021enlightengan} 
addresses enhancement of low-light images and ToDayGAN~\cite{anoosheh2019night} exploits night-to-day image translation for retrieval-based localization. Recently, ForkGAN~\cite{zheng2020forkgan} presents reasonable translation outputs in rainy night~$\rightarrow$~day task by adopting a common encoding space from a different domain.

\paragraph{Uncertainty-aware learning}
The uncertainty-aware learning considers modeling the uncertainty of the data or predictions by the network. Specifically, we consider heteroscedastic aleatoric uncertainty~\cite{kendall2017uncertainties} that captures heteroscedastic noise inherent in the observations. Recently, the heteroscedastic regression is exploited in several vision tasks such as depth estimation~\cite{eldesokey2020uncertainty} or 3D reconstruction from a single 2D image~\cite{wu2020unsupervised}. This approach is useful when specific regions of observation have higher noise than other parts. There are regions with blur and reflections in rainy night images, thus these regions can be interpreted as having higher uncertainty than others. Therefore, we apply the heteroscedastic regression to minimize difference between an original image and a cyclic reconstructed one.

\begin{figure}[!t]
\centering \includegraphics[width=\linewidth]{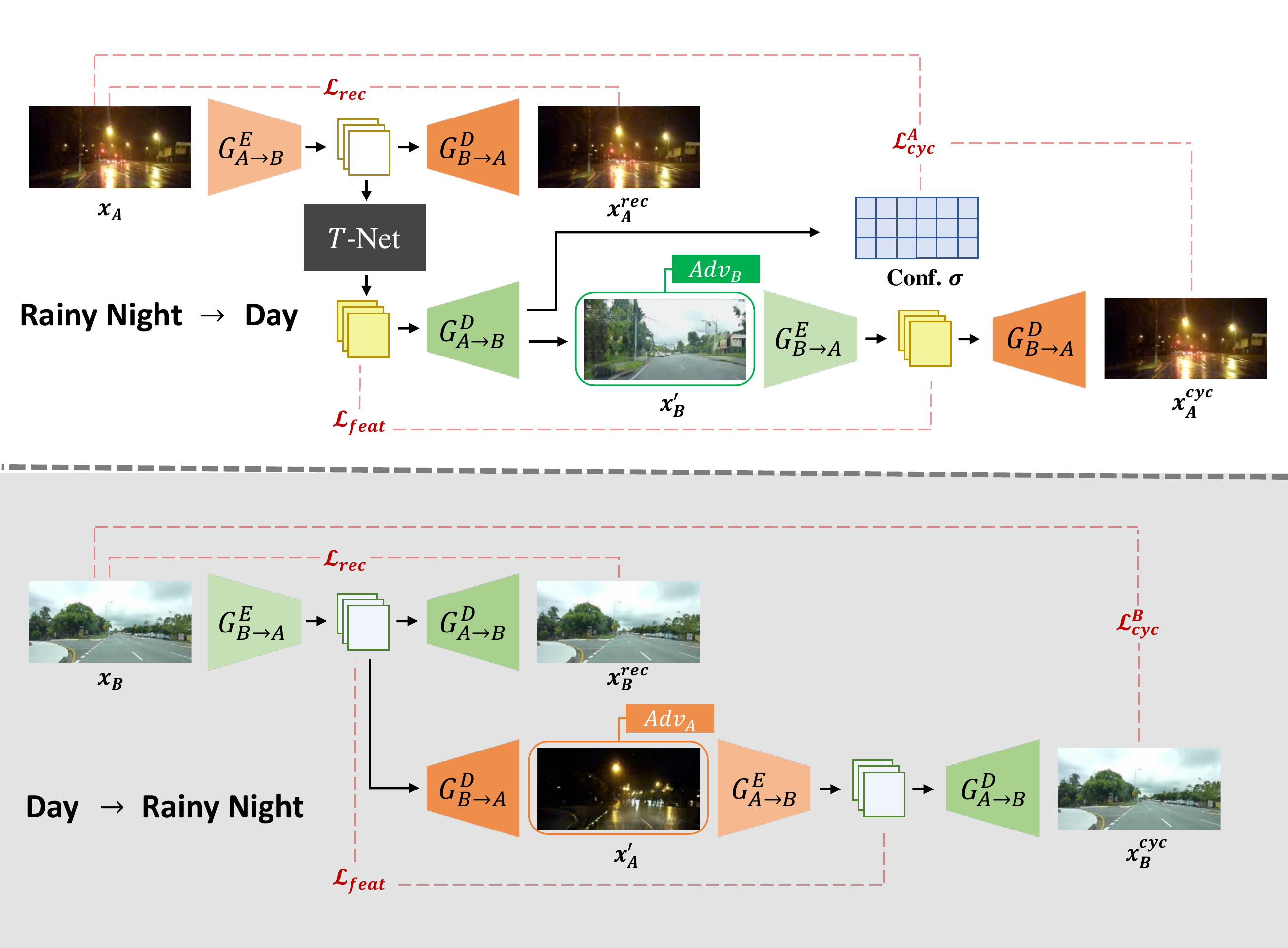}
\caption{Overview of our model. On the upper side is procedure of rainy night $\rightarrow$ day and the opposite is that of day $\rightarrow$ rainy night.} 
\label{fig:f2}
\end{figure}  

\section{Proposed method} \label{sec:method}
This section describes our proposed framework to address adverse weather image translation in detail, by first presenting a model overview and describing the proposed loss functions. 
\subsection{Asymmetric architecture}
Let $x_{\mathcal{A}}\in\mathcal{A}$ and $x_{\mathcal{B}}\in\mathcal{B}$ denote images from adverse domain $\mathcal{A}$ (rainy night) and standard domain $\mathcal{B}$ (daytime), respectively.
As shown in Fig.~\ref{fig:f2}, there are two generators consisting of an encoder and a decoder, i.e., $G_{\mathcal{A\rightarrow B}}=\{G^{E}_{\mathcal{A\rightarrow B}},G^{D}_{\mathcal{A\rightarrow B}}\}$ which converts domain $\mathcal{A}$ to $\mathcal{B}$ ($\mathcal{A}\rightarrow\mathcal{B}$) and $G_{\mathcal{B\rightarrow A}}=\{G^{E}_{\mathcal{B\rightarrow A}},G^{D}_{\mathcal{B\rightarrow A}}\}$ which converts domain $\mathcal{B}$ to $\mathcal{A}$ ($\mathcal{B}\rightarrow\mathcal{A}$).
The goal of adverse weather translation is to synthesize successfully edited image $x'_{\mathcal{B}}$ from $x_{\mathcal{A}}$ with the generator $G_{\mathcal{A\rightarrow B}}$. 
Most of CycleGAN-based models adopt cyclic translation procedure ($\mathcal{A}\rightarrow\mathcal{B}\rightarrow\mathcal{A}$) to exploit cycle-consistency loss and also include a symmetrical opposite translation ($\mathcal{B}\rightarrow\mathcal{A}\rightarrow\mathcal{B}$) for stable and balanced optimization. 
ForkGAN~\cite{zheng2020forkgan} presents a constraint to enforce encoded intermediate features to be domain-invariant. 
While maintaining these spirits, but unlike existing methods, we propose a novel asymmetric framework for image translation. 
The reason why we do {\bf NOT} adopt symmetric procedure is quite intuitive. 
Suppose the encoder   $G^{E}_{\mathcal{A\rightarrow B}}$ could acquire domain-invariant feature. 
With the feature, reconstructed image $x^{rec}_\mathcal{A}$ and transferred image $x'_\mathcal{B}$ are synthesized by $G^{D}_{\mathcal{B\rightarrow A}}$ and    $G^{D}_{\mathcal{A\rightarrow B}}$ respectively, and then $G_{\mathcal{B\rightarrow A}}$ generates cyclic image $x^{cyc}_\mathcal{A}$ based on $x'_{B}$. 
In training phase, the differences from original image $x_{\mathcal{A}}$, i.e.,  reconstruction loss ($x_{\mathcal{A}}$ vs. $x^{rec}_\mathcal{A}$) and cycle-consistency loss ($x_{\mathcal{A}}$ vs. $x^{cyc}_\mathcal{A}$), are included in objectives. 
However, if the encoder extracts "truly" domain-invariant features, it is impossible to reconstruct the original adverse weather image perfectly.
Because the negative domain-specific characteristics (\eg, artifacts and reflections) are removed in the feature. 
Therefore, there is a dilemma that the feature from the adverse domain image should preserve several domain-specific characteristics for reconstruction but exclude them for translation.   
To address this issue, we insert an additional transfer network (${T}$-net) which consists of several residual blocks~\cite{he2016deep} only inside of generator $G_{\mathcal{A\rightarrow B}}$ to acquire an enhanced and disentangled feature for domain translation.      
Consequently, as shown in Fig.~\ref{fig:f2}, two domain translation functions ($f_{\mathcal{A}\rightarrow\mathcal{B}}$ and $f_{\mathcal{B}\rightarrow\mathcal{A}}$) are not symmetrical, i.e.,
\begin{align}
    x'_{\mathcal{B}}=f_{\mathcal{A}\rightarrow\mathcal{B}}(x_{\mathcal{A}})&= G^{D}_{\mathcal{A\rightarrow B}}({T}(G^{E}_{\mathcal{A\rightarrow B}}(x_{\mathcal{A}}))) , \label{eq:1}\\
    x'_{\mathcal{A}}=f_{\mathcal{B}\rightarrow\mathcal{A}}(x_{\mathcal{B}})&=
    G^{D}_{\mathcal{B\rightarrow A}}(G^{E}_{\mathcal{B\rightarrow A}}(x_{\mathcal{B}}))), \label{eq:2} 
\end{align}
and the reconstruction procedures ($f_{\mathcal{A}\rightarrow\mathcal{A}}$ and $f_{\mathcal{B}\rightarrow\mathcal{B}}$) can be expressed as 
\begin{align}
    x^{rec}_{\mathcal{A}}=f_{\mathcal{A}\rightarrow\mathcal{A}}(x_{\mathcal{A}})&= G^{D}_{\mathcal{B\rightarrow A}}(G^{E}_{\mathcal{A\rightarrow B}}(x_{\mathcal{A}}))) , \label{eq:3}\\
    x^{rec}_{\mathcal{B}}=f_{\mathcal{B}\rightarrow\mathcal{B}}(x_{\mathcal{B}})&=
    G^{D}_{\mathcal{A\rightarrow B}}(G^{E}_{\mathcal{B\rightarrow A}}(x_{\mathcal{B}}))). \label{eq:4} 
\end{align}
 Lastly, cyclic operations ($f_{\mathcal{A}\rightarrow\mathcal{B}\rightarrow\mathcal{A}}$ and $f_{\mathcal{B}\rightarrow\mathcal{A}\rightarrow\mathcal{B}}$) is represented as
 \begin{align}
    x^{cyc}_{\mathcal{A}}=f_{\mathcal{A}\rightarrow\mathcal{B}\rightarrow\mathcal{A}}(x_{\mathcal{A}})&= G^{D}_{\mathcal{B\rightarrow A}}(G^{E}_{\mathcal{B\rightarrow A}}(x'_{\mathcal{B}}))) , \label{eq:5}\\
    x^{cyc}_{\mathcal{B}}=f_{\mathcal{B}\rightarrow\mathcal{A}\rightarrow\mathcal{B}}(x_{\mathcal{B}})&=
    G^{D}_{\mathcal{A\rightarrow B}}(G^{E}_{\mathcal{A\rightarrow B}}(x'_{\mathcal{A}}))). \label{eq:6}
\end{align}
We use a pixel-level $\ell_1$ loss to assure the reconstruction ability for each domain, i.e., 
\begin{equation} \label{eq:7}
    \mathcal{L}_{rec}=\mathbb{E}_{x_{\mathcal{A}}}[\lVert{x_{\mathcal{A}}-x^{rec}_{\mathcal{A}}}\rVert_1]+\mathbb{E}_{x_{\mathcal{B}}}[\lVert{x_{\mathcal{B}}-x^{rec}_{\mathcal{B}}}\rVert_1].
\end{equation}
In addition, the extracted domain invariant feature by each encoder should be disentangled from the domain specific feature. However, encoded feature $G^{E}_{\mathcal{A\rightarrow B}}(x_{\mathcal{A}})$ could not be perfect domain-invariant feature. Because the information of domain-variant artifacts such as rain drop or reflection that should be removed for translation still remains in encoded feature for reconstruction.       
${T}$-net plays an important role in making the encoded feature be more informative and disentangled from domain-variant information thus alleviates burden of the adverse domain encoder $G^{E}_{\mathcal{A\rightarrow B}}$.  
To this end, we present asymmetric feature matching loss for disentanglement, where the loss penalizes the difference between the encoded feature of input image and that of transferred image by different encoders respectively, i.e., 

\begin{equation} \label{eq:8}
    \mathcal{L}_{feat}=\mathbb{E}_{x_{\mathcal{A}}}[\lVert{{T}(G^{E}_{\mathcal{A\rightarrow B}}(x_{\mathcal{A}}))-G^{E}_{\mathcal{B\rightarrow A}}(x'_{\mathcal{B}})}\rVert_1]+
    \mathbb{E}_{x_{\mathcal{B}}}[\lVert{(G^{E}_{\mathcal{B\rightarrow A}}(x_{\mathcal{B}}))-G^{E}_{\mathcal{A\rightarrow B}}(x'_{\mathcal{A}})}\rVert_1],
\end{equation}

\noindent here, note that the extracted feature from adverse domain image $x_{\mathcal{A}}$ is compared after passing through ${T}$-net.    

\subsection{Uncertainty-aware cyclic loss}
Although the procedures of reconstruction ($\mathcal{A\rightarrow B}$) and  translation ($\mathcal{B\rightarrow A}$) are separated by utilizing ${T}$-net, there is still ambiguity in cyclic reconstruction process of adverse domain image ($\mathcal{A}\rightarrow\mathcal{B}\rightarrow\mathcal{A}$). 
Because the domain-specific characteristics of $\mathcal{A}$ no longer need to be preserved in transformed image $x'_\mathcal{B}$ exactly. In other words, it is not necessary for the model to accurately reconstruct artifacts or reflections by raindrops.     
As a result, applying the cycle-consistency loss uniformly to all regions can lead to a trivial solution and poor convergence of optimization. 
Motivated by uncertainty modeling~\cite{kendall2017uncertainties}, we modify $G^{D}_{\mathcal{A\rightarrow B}}$ to generate not only the transformed image $x'_{\mathcal{B}}$ but also a confidence map (uncertainty map) $\sigma\in\mathbb{R}_{+}^{H\times{W}}$. The confidence map $\sigma$, which is estimated during training without supervision, models the uncertainty of the model, specifically aleatoric uncertainty. We propose uncertainty-aware cycle-consistency loss to address regional difference of an input image from adverse domain $\mathcal{A}$ With $\sigma$, i.e.,

\begin{equation}
    \mathcal{L}_{cyc}^\mathcal{A} = \frac{1}{HW} \sum_{i=1}^{W}\sum_{j=1}^{H}{ \frac{1}{2}\sigma^{-2}_{ij}{\lVert{x_{\mathcal{A}_{ij}}-x^{cyc}_{\mathcal{A}_{ij}}}\rVert}_{1}+\frac{1}{2}\log\sigma^{2}_{ij}},
\label{eq:eq9}
\end{equation}

\noindent where $x_{\mathcal{A}_{ij}}$ and $x^{cyc}_{\mathcal{A}_{ij}}$ denote the pixel intensity at location $(i,j)$ of $x_{\mathcal{A}}$ and $x^{cyc}_{\mathcal{A}}$ respectively and $\sigma_{ij}$ means the estimation of uncertainty at $(i,j)$. Eq.~(\ref{eq:eq9}) can be interpreted that the regions with large uncertainty are less affected by pixel-level difference, but penalized by the increased regularization term.   
The cyclic loss for normal domain $\mathcal{B}$ is a pixel-level $\ell_1$ loss between $x_{\mathcal{B}}$ and $x^{cyc}_{\mathcal{B}}$ generally used in CycleGAN-based methods, i.e.,

\begin{equation}
    \mathcal{L}_{cyc}^\mathcal{B} = \mathbb{E}_{x_{\mathcal{B}}}[\lVert{x_{\mathcal{B}}-x^{cyc}_{\mathcal{B}}}\rVert_1],
\label{eq:eq10}
\end{equation}

\noindent hence the overall cyclic loss is calculated as,
\begin{equation}
    \mathcal{L}_{cyc} = \mathcal{L}_{cyc}^\mathcal{A} + \mathcal{L}_{cyc}^\mathcal{B}.
\label{eq:eq11}
\end{equation}
\subsection{Model objective}
In addition to the aforementioned loss functions, we exploit an adversarial training to guarantee visually realistic output through domain-specific discriminators which distinguish the real and fake image. In detail, we adopt LSGAN~\cite{mao2017least} loss to minimize the discrepancy between the distribution of real and that of the translated image. Therefore, the adversarial loss of generator and discriminator related to domain $\mathcal{B}$  can be described as

\begin{align}
    \mathcal{L}^{\mathcal{B}}_{D_{adv}}&=\frac{1}{2}\mathbb{E}_{x_\mathcal{B}}[(D_{\mathcal{B}}(x_\mathcal{B})-1)^{2}]
    + \frac{1}{2}\mathbb{E}_{x'_\mathcal{B}}[(D_{\mathcal{B}}(x'_\mathcal{B}))^{2}], \\
    \mathcal{L}^{\mathcal{B}}_{G_{adv}}&=\frac{1}{2}\mathbb{E}_{x'_\mathcal{B}}[(D_{\mathcal{B}}(x'_\mathcal{B})-1)^{2}],
\end{align}

\noindent where, $D_{\mathcal{B}}$ denotes the discriminator of domain $\mathcal{B}$. Note that also the adversarial losses related to domain 
${\mathcal{A}}$ , i.e., ${L}^{\mathcal{A}}_{D_{adv}}$ and ${L}^{\mathcal{A}}_{G_{adv}}$, are obtained by $D_{\mathcal{A}}$ in the same way as $D_{\mathcal{B}}$ but we omit the details of them.    

Finally, the full objective of our model is formulated as

\begin{align}
    \mathcal{L}_{D} &= \mathcal{L}^{\mathcal{A}}_{D_{adv}}+\mathcal{L}^{\mathcal{B}}_{D_{adv}},  
    \label{eq:eq14}
    \\ 
        \mathcal{L}_{G} &= \mathcal{L}^{\mathcal{A}}_{G_{adv}}+\mathcal{L}^{\mathcal{B}}_{G_{adv}} + 
        \lambda_{rec}\mathcal{L}_{rec}+\lambda_{feat}\mathcal{L}_{feat}+\lambda_{cyc}\mathcal{L}_{cyc},
        \label{eq:eq15}
\end{align}

\noindent where $\lambda_{rec}$, $\lambda_{feat}$ and $\lambda_{cyc}$ are hyper-parameters which modulate the relative importance of the terms. 


\section{Experiments}
In this section, we first explain our experimental setup and then present qualitative and quantitative comparisons of ours with the state-of-the-art methods, i.e., CycleGAN~~\cite{Zhu_2017}, 
UNIT~\cite{liu2017unsupervised}, ToDayGAN~\cite{anoosheh2019night}
and ForkGAN~\cite{zheng2020forkgan}. For each method, we use the official implementations provided by the authors.  Finally, we demonstrate the effectiveness of each element of the proposed method through an ablation study. 

\subsection{Experimental setup}
For experiments, we use two datasets, i.e., Alderley Day/Night Dataset (Alderley)~\cite{6224623} and Berkeley DeepDrive dataset (BDD100K)~\cite{yu2018bdd100k}. Alderley consists of images of two domains, rainy night and daytime. It was collected while driving the same route in each weather environment. A lot of images taken from the rainy nights have reflections or artifacts by raindrops and the regions without light are difficult to be recognized. On the contrary, almost daytime images are clean and objects in them are plainly visible. We evaluate models' capabilities of translating adverse weather image with Alderley. BDD100K contains 100,000 high resolution images of the urban roads for autonomous driving. 
There are 10K images in the package of BDD (BDD10K), have its label for semantic segmentation. We use them to estimate the performance of pretrained segmentation model~\cite{zhao2017pyramid} given translated images. 
In experiments, the resolution of all input and output images is $256~\times~512$ and we adopt Adam~\cite{kingma2014adam} solver with $\beta_1=0.5$ and $\beta_2=0.999$. Coefficients of the full objective in Eq.~(\ref{eq:eq15}) are set to $\lambda_{feat}=1$ and $\lambda_{rec}=\lambda_{cyc}=10$ and the learning rate is set to 0.0002.                           
\begin{figure}[!t]
\centering \includegraphics[width=\linewidth]{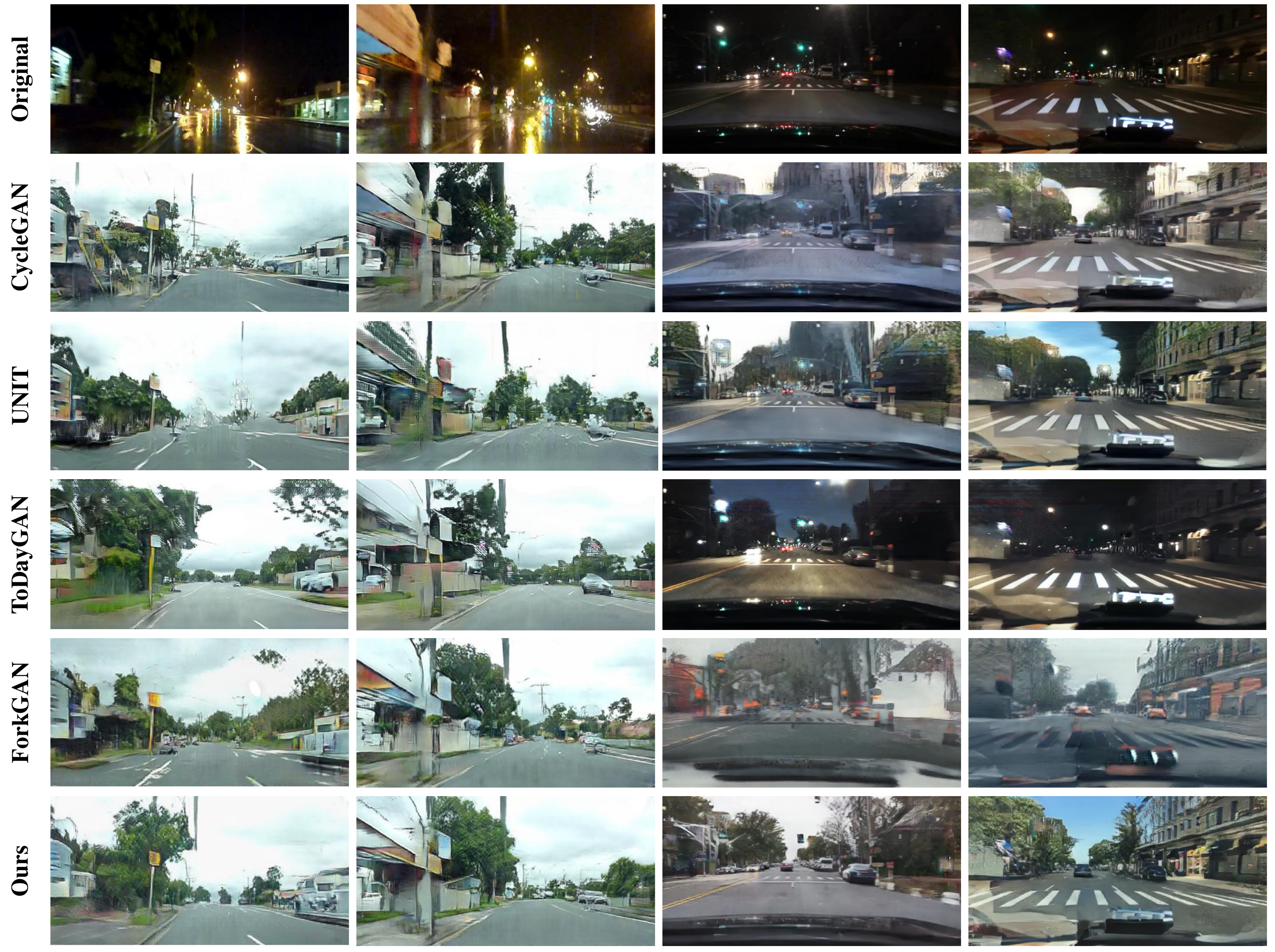}
\caption{Qualitative comparison with unsupervised image-to-image translation models for adverse weather enhancement, i.e., CycleGAN~\cite{Zhu_2017}, UNIT~\cite{liu2017unsupervised}, ToDayGAN~\cite{anoosheh2019night}, ForkGAN~\cite{zheng2020forkgan} and our model. Left two columns: Alderley (rainy night$\rightarrow$day) and right two columns : BDD100K (night$\rightarrow$day).  \textbf{Please zoom in to see more details.}}
\label{fig:f3}
\end{figure}  
\subsection{Qualitative result} \label{qual}
We first present the results of qualitative comparison with three the-state-of-art methods in adverse weather image translation, i.e., CycleGAN~\cite{Zhu_2017}, UNIT~\cite{liu2017unsupervised}, ToDayGAN~\cite{anoosheh2019night} and ForkGAN~\cite{zheng2020forkgan}. The results are shown in Fig.~\ref{fig:f3}. Source image from the adverse domain is placed in uppermost of each column, and each row corresponds to translation outputs from each method. The left two columns represent qualitative translation results from Alderley (rainy night $\rightarrow$ day) and the right two columns represent that from BDD100K (night $\rightarrow$ day).   
Although CycleGAN performs editing properly on the regions where the objects are clearly visible 
, the translated results for the dark or blurry regions show inferior visual quality. Although   TodayGAN and UNIT present improved editing ability in experiments on the Alderley dataset compared to CycleGAN, but they generate several artifacts and the outputs are not transformed properly when conducting using BDD100K. Similarly, ForkGAN which exploits a common encoding space of two domian produces overall blurry images in the experiments on BDD100K. This is because they are not converged well when training using BDD100K, which has huge diversity. As shown in Fig.~\ref{fig:f3}, our model can successfully conduct adverse weather translation with both datasets. It outputs visually superior results compared to other methods in most regions including dark and blurry areas. In addition, existing objects are well preserved in the transformed image.      
\subsection{Quantitative result}
In this section, we report quantitative result with two metrics, i.e., FID (Fréchet Inception Distance) score~\cite{heusel2017gans} computed with the extracted feature of Inception network~\cite{szegedy2015going} and mIOU (mean of class-wise Intersection over Union) obtained by the result of pretrained semantic segmentation model~\cite{zhao2017pyramid}.   

\paragraph{FID score.}
We first present the FID score which is commonly utilized as a metric of GAN models for evaluating visual quality. We evaluate the FID score by comparing two sets of images, i.e., real images (day) vs. transformed fake images (rainy night $\rightarrow$ day), and each set contains 1,000 images in the test set. The results are listed in Table~\ref{tb:tb1}, where "real." denotes a set of adverse domain images (rainy night or night). Our model outperforms other methods in both Alderley and BDD100K by a large gap. We analyze the results of variants of our method (Ours~w.o.~$T$ and Ours~w.o.~$un$) in Sec.~\ref{sec:abl}.       

\begin{table}
\caption{FID (Fréchet Inception Distance) scores. Lower is better.}
\begin{center}
{\small
\begin{tabular}{p{0.2cm}p{0.8cm}p{0.8cm}p{0.8cm}p{0.8cm}p{0.8cm}p{0.8cm}p{1.1cm}p{1.0cm}p{0.8cm}}
\toprule
             && Cycle-GAN & UNIT    & ToDay-GAN     & Fork-GAN   & Ours & Ours w.o. $un.$  & Ours w.o. $T$ &  real. \\
\midrule
Alderley  &&102.4 & 88.5 &98.5 &75.8 &\textbf{65.2} & 76.4 & 83.3 & 189.2  \\
BDD100K &&53.1 & 62.4 &78.9 &63.0 &\textbf{38.6} & 42.5 & 55.1 & 98.3\\     
\bottomrule
\end{tabular}
}
\end{center}
\label{tb:tb1}
\end{table}

\begin{table}
\caption{Semantic segmentation results (mIOU) on translated images conducted by PSPNet~\cite{zhao2017pyramid} with ResNet-101~\cite{he2016deep} as a backbone. Numbers indicate the percentage of mIOU.}
\begin{center}
{\small
\begin{tabular}{p{0.2cm}p{0.8cm}p{0.8cm}p{0.8cm}p{0.8cm}p{0.8cm}p{0.8cm}p{1.1cm}p{1.0cm}p{0.8cm}}
\toprule
            && Cycle-GAN  & UNIT  & ToDay-GAN     & Fork-GAN   & Ours & Ours w.o. $un.$  & Ours w.o. $T$ &  real. \\
\midrule
BDD10K  &&15.48 &13.18 &8.91 &10.15 &\textbf{18.57} &17.62 & 14.08 & 12.33 \\
\bottomrule
\end{tabular}
}
\end{center}
\label{tb:tb2}
\end{table}

\paragraph{Semantic segmentation}
To measure the effect of domain translation on the performance of other computer vision models, we report semantic segmentation results of synthesized images. For semantic segmentation, we exploit PSPNet~\cite{zhao2017pyramid} with ResNet-101~\cite{he2016deep} as a backbone which pretrained on Cityscapes dataset~\cite{7780719}. For domain transfer, night images of the BDD10K validation set that has a ground-truth segmentation label corresponding to each image are used as input. After the image translation, we compute mIOU using PSPNet and the results are presented in Table.~\ref{tb:tb2}. In the case of ToDayGAN and ForkGAN, the segmentation performance are rather decrease compared to before translation. CycleGAN and UNIT slightly enhance the segmentation result. Our model brings further improvement on semantic segmentation task compared to other methods. 

\subsection{Ablation study} \label{sec:abl}
\begin{figure}[!t]
\centering \includegraphics[width=\linewidth]{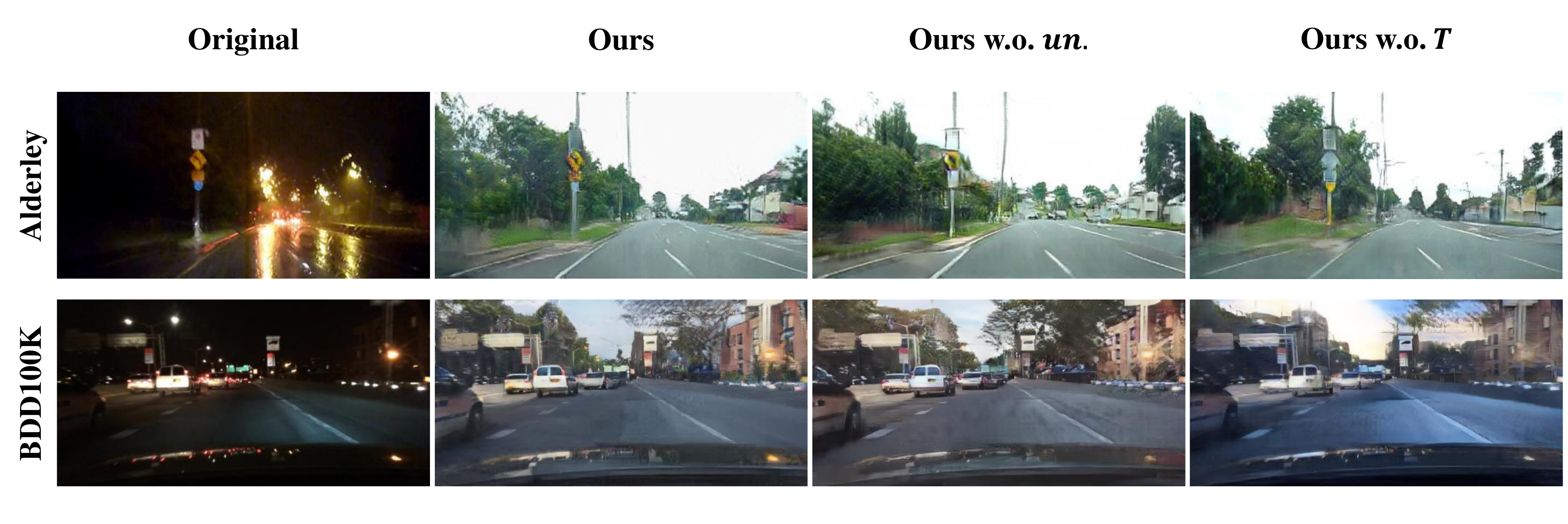}
\caption{Qualitative results of ablation study. \textbf{Please zoom in to see more details.}}
\label{fig:f4}
\end{figure}

To demonstrate the effectiveness of our method, we present qualitative and quantitative results by excluding key components one by one. We compare the proposed model with two different versions, i.e., (i) Ours w.o. $un.$ : excluding uncertainty-based cyclic loss $\mathcal{L}_{cyc}^\mathcal{A}$, here we adopt existing cycle-consistency loss in both translation directions ($\mathcal{A}\rightarrow\mathcal{B}\rightarrow\mathcal{A}$ and $\mathcal{B}\rightarrow\mathcal{A}\rightarrow\mathcal{B}$) and (ii) Ours w.o. $T$: removing  $T$-net in our generator $G_{\mathcal{A\rightarrow B}}$. Here, we also remove $\mathcal{L}_{cyc}^\mathcal{A}$ because uncertainty map is calculated from the feature passing $T$-net.  Fig.~\ref{fig:f4} presents qualitative results of each variant and the quantitative results are list in Table.~\ref{tb:tb1} and Table.~\ref{tb:tb2}. In the domain with low uncertainty (BDD100K), the visual degradation when discarding uncertainty-aware loss is slight but it is enlarged in the domain with high uncertainty due to artifacts or reflection by raindrop (Alderley). $T$-net helps to improve the overall visual quality of the translated images as well as to preserve existing objects by enhancing feature disentanglement as shown in Fig.~\ref{fig:f4}, i.e., Ours vs. Ours w.o $T$. In addition, the quantitative metrics also support this analysis. When uncertainty-aware loss is excluded, there is a larger difference in the FID score on Alderley than BDD100K. Removal of $T$-net causes noticeable degradation of FID score on both datasets. The segmentation performance (mIOU) of transferred images also decreased as the proposed elements are removed one by one.          

\subsection{Visualization of uncertainty} \label{sec:vis}
\begin{figure}[!t]
\centering \includegraphics[width=\linewidth]{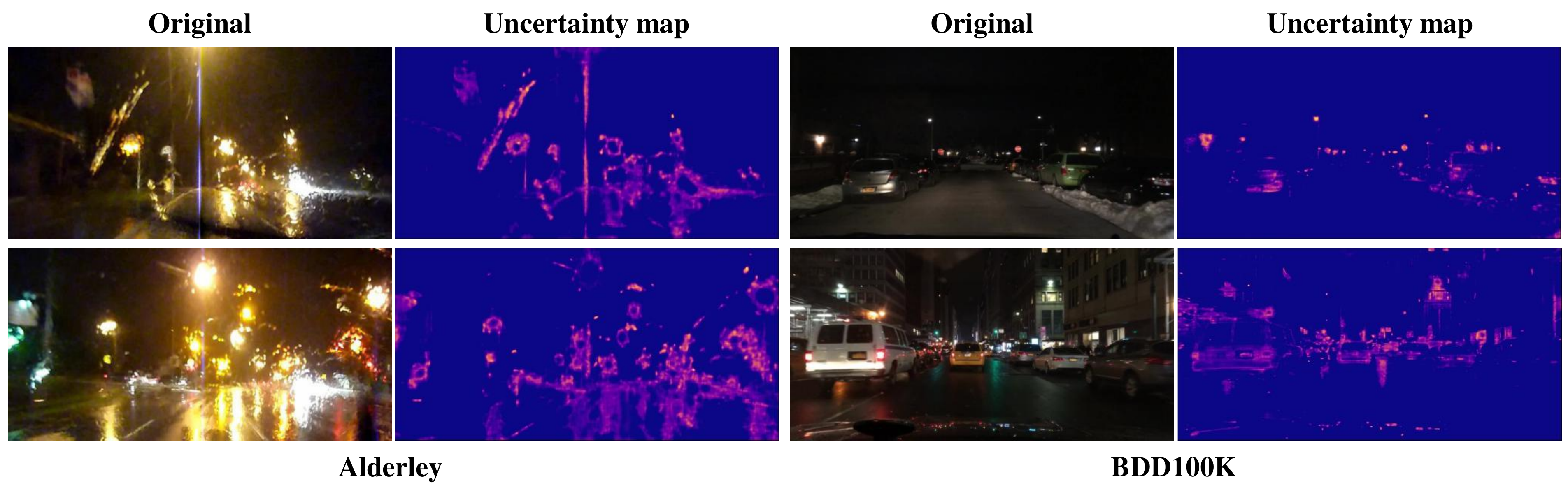}
\caption{Visualization results of the predicted uncertainty map $\sigma$ (conf.) }
\label{fig:f5}
\end{figure}

Fig.~\ref{fig:f5} shows visualization result of the predicted uncertainty map to examine whether it captures the uncertainty of regions correctly. Here, the yellow or purple areas in uncertainty map mean regions with high uncertainty. Our purpose of introducing uncertainty map and $\mathcal{L}_{cyc}^\mathcal{A}$ is that the regions with high uncertainty are less penalized by $\ell_1$ loss term but more by regularization term in cyclic reconstruction procedure (Eq.~\ref{eq:8}). As we expected, the adverse parts, \eg, glare around street lamp, reflections of wet road, rain drops and wiper mark in Alderley dataset have high uncertainty values (left in Fig.~\ref{fig:f5}). Although there are some regions with high uncertainty in BDD100K such as reflection of light (right in Fig.~\ref{fig:f5}), the overall uncertainty of Alderley is higher than BDD100K because Alderley consists of more adverse domain (rainy night) than BDD100K (night). The translated result of each input can be found in qualitative results of Sec.~\ref{qual} and supplementary file. 
The advantage of our method is that it can adaptively learn and handle uncertainty of each dataset without any supervision.                   

\section{Conclusion}
In this paper, we introduced the asymmetric and uncertainty-aware GAN model to address adverse weather image translation. 
We separated reconstruction and translation by adopting a feature transfer network ($T$-net) that enhances disentanglement of the encoded feature of adverse domain image. In addition, we analyzed the limitation of application cycle-consistency loss in adverse domain transfer and propose the uncertainty-based cycle-consistency loss by estimating the confidence map in the generator. The superiority of our method is demonstrated by comparisons with state-of-the-art methods through qualitative and quantitative studies. In the future, we will extend our method to multi-modal translation and jointly exploiting attention-based techniques.      
 
\paragraph{Acknowledgement} This research was supported by Deep Machine Lab (Q2109331)

\clearpage

\section{Appendix}
In this section, we supplement additional analysis and experimental results that are not presented in the main paper. We first describe the details of our architecture for reproducibility (Sec.~\ref{sec:sec6.1}) and then analyze the effectiveness of the uncertainty-aware cycle consistency loss (Sec.~\ref{sec:sec6.2}). In addition, we conduct broader analysis of our method, i.e., experiments on higher resolution ($512\times1024$ and failure case (Sec.~\ref{sec:sec6.5}). Finally, we present additional comparison results (Sec.~\ref{sec:sec6.3}) and extra qualitative results of our model (Sec.~\ref{sec:sec6.4}).
\subsection{Implementation}
\begin{figure}[!t]
\centering \includegraphics[width=\linewidth]{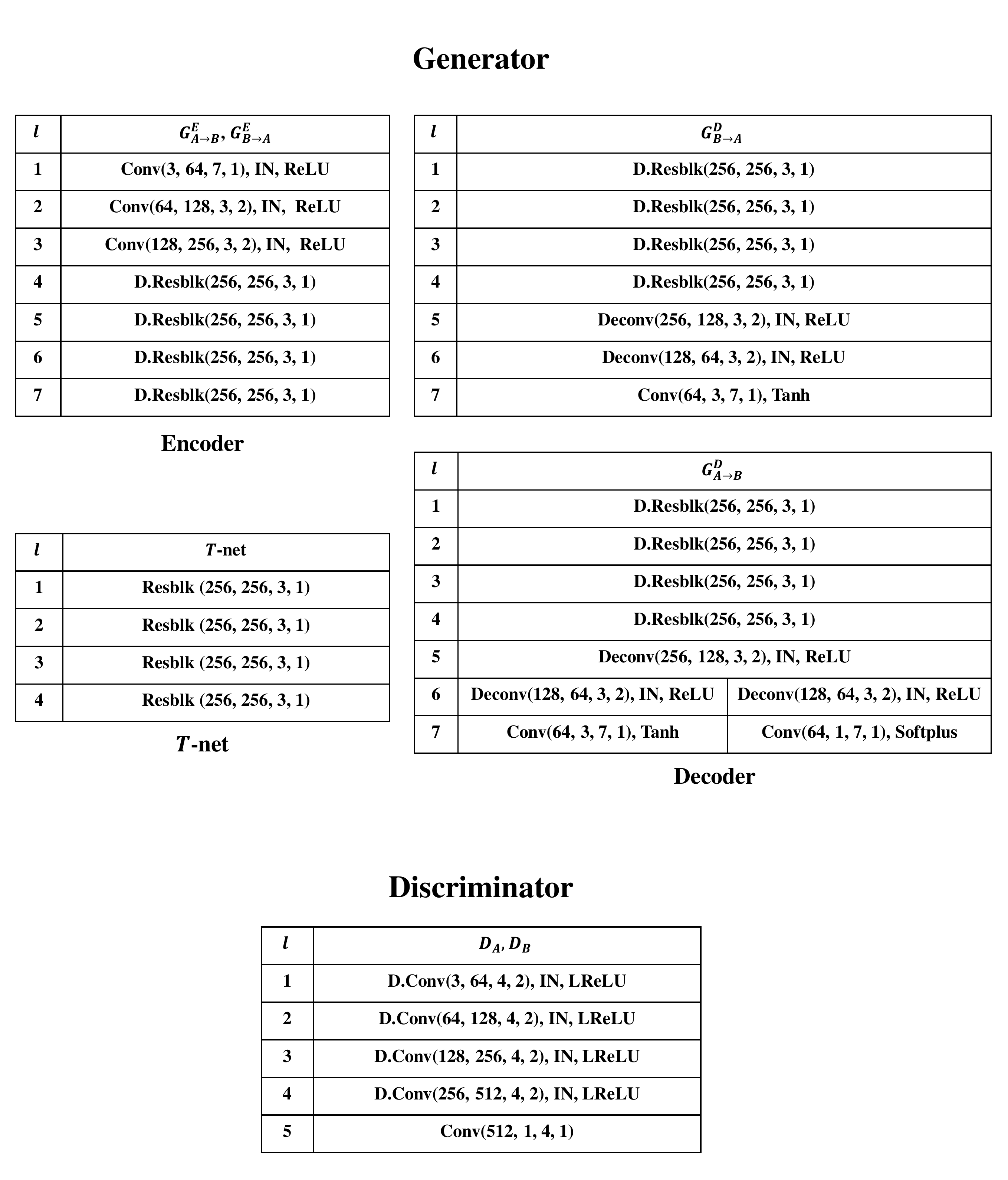}
\caption{Details of proposed modules. Conv, Resblk, D.Resblk, Deconv denotes convolutional layer, residual block, residual block with dilated convolution and transposed convolution respectively. ($c_{in}, c_{out}, k, s$) denotes input channels, output channels, kernel size, stride respectively.} 
\label{fig:sf1}
\end{figure}

We report details of each module of our model and figures are depicted in Fig.~\ref{fig:sf1}. In the following, we explain each module.   

\label{sec:sec6.1}
\paragraph{Encoder}
The encoders of two domains $\{G^{E}_{\mathcal{A\rightarrow B}}, G^{E}_{\mathcal{B\rightarrow A}}\}$ have same network architecture. They consist of three convolutional layers and four residual blocks~\cite{he2016deep} with dilated convolution~\cite{yu2015multi} (D.Resblk). Therefore, an input image, i.e., $x_{\mathcal{A},\mathcal{B}}\in\mathbb{R}^{256\times{512}\times{3}}$ , is converted to encoded feature with the output size in  ${\mathbb{R}^{ 64\times{128}\times{256}}}$. In addition, we utilize Instance Normalization~\cite{ulyanov2016instance} (IN). in all layers of the encoder. 

\paragraph{$T$-net}
As mentioned in main text, feature transfer network ($T$-net) is inserted in $G_{\mathcal{A\rightarrow B}}$. It consists of four residual blocks (Resblk) thus the size of input and output is same as in ${\mathbb{R}^{64\times{128}\times{256}}}$.   

\paragraph{Decoder}
The two decoders $\{G^{D}_{\mathcal{A\rightarrow B}}, G^{D}_{\mathcal{B\rightarrow A}}\}$ have same structure except for the last two layers. They have a symmetrical structure with the encoders, thus the input feature $\in\mathbb{R}^{64\times{128}\times{256}}$ is transformed to RGB output image $\in\mathbb{R}^{256\times{512}\times{3}}$ by transposed convolution (Deconv). Unlike $G^{D}_{\mathcal{B\rightarrow A}}$, $G^{D}_{\mathcal{A\rightarrow B}}$ has an additional branch that generates the uncertainty map $\sigma$. With the mid-feature $\in\mathbb{R}^{128\times{256}\times{128}}$ in decoder, the branch outputs the uncertainty map$\in\mathbb{R_{+}}^{256\times{512}}$ by including Softplus in the last layer.

\paragraph{Discriminator}
The discriminators \{$D_{\mathcal{A}}, D_{\mathcal{B}}$\} have the form of multi-scale ~\cite{wang2018high} and PatchGAN~\cite{Isola_2017_CVPR} discriminators. The resolution of the output activations are in ${\mathbb{R}^{16\times{32}}}$ and ${\mathbb{R}^{8\times{16}}}$. As similar with the generators, we use Instance Normalization in all layers of each discriminator except for the last layer.

\begin{figure}[!t]
\centering \includegraphics[width=\linewidth]{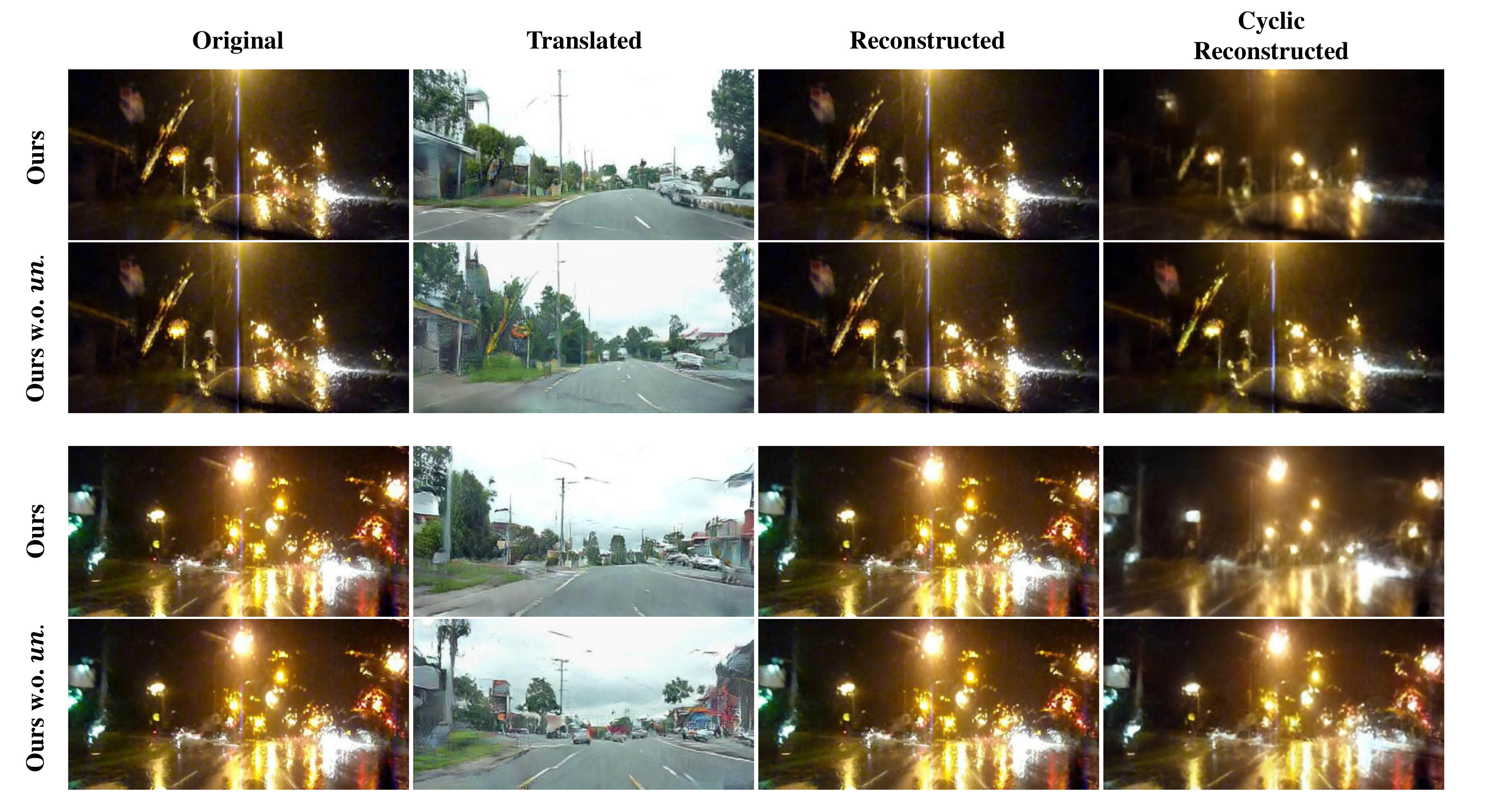}
\caption{Experiments results of the variants of our model, i.e., the uncertainty-aware loss is used or not.} 
\label{fig:sf2}
\end{figure}  

\subsection{Analysis of uncertainty-aware cyclic loss}
\label{sec:sec6.2}
To demonstrate the effectiveness of the uncertainty-aware cycle consistency loss $\mathcal{L}_{cyc}^\mathcal{A}$, we analyze the role of the loss in training. We compare the translated images ($\mathcal{A\rightarrow B}$), the reconstructed images ($\mathcal{A\rightarrow A}$) and the cyclic reconstructed images ($\mathcal{A\rightarrow B \rightarrow A}$) of two variants of our model, i.e., with $\mathcal{L}_{cyc}^\mathcal{A}$ (Ours) and without $\mathcal{L}_{cyc}^\mathcal{A}$ (Ours w.o. $un.$). In the latter case, we use the standard cycle consistency loss~\cite{Zhu_2017}. The results are presented in Fig.~\ref{fig:sf2}. As mentioned in the main paper, the cyclic reconstructed image is not obliged to possess artifacts same as that of the original if the disentanglement is conducted successfully. As shown in Fig.~\ref{fig:sf2}, the cyclic reconstructed image when $\mathcal{L}_{cyc}^\mathcal{A}$ is not in use has unnecessary artifacts or reflections. As a result, some of artifacts also appear in the transferred day image. However when using $\mathcal{L}_{cyc}^\mathcal{A}$, the problem is alleviated because the regions with artifacts or reflections have less confidence and thus they are removed clearly in the converted day image.

\subsection{Broader analysis}
\label{sec:sec6.5}

\begin{figure}[!t]
\centering \includegraphics[width=\linewidth]{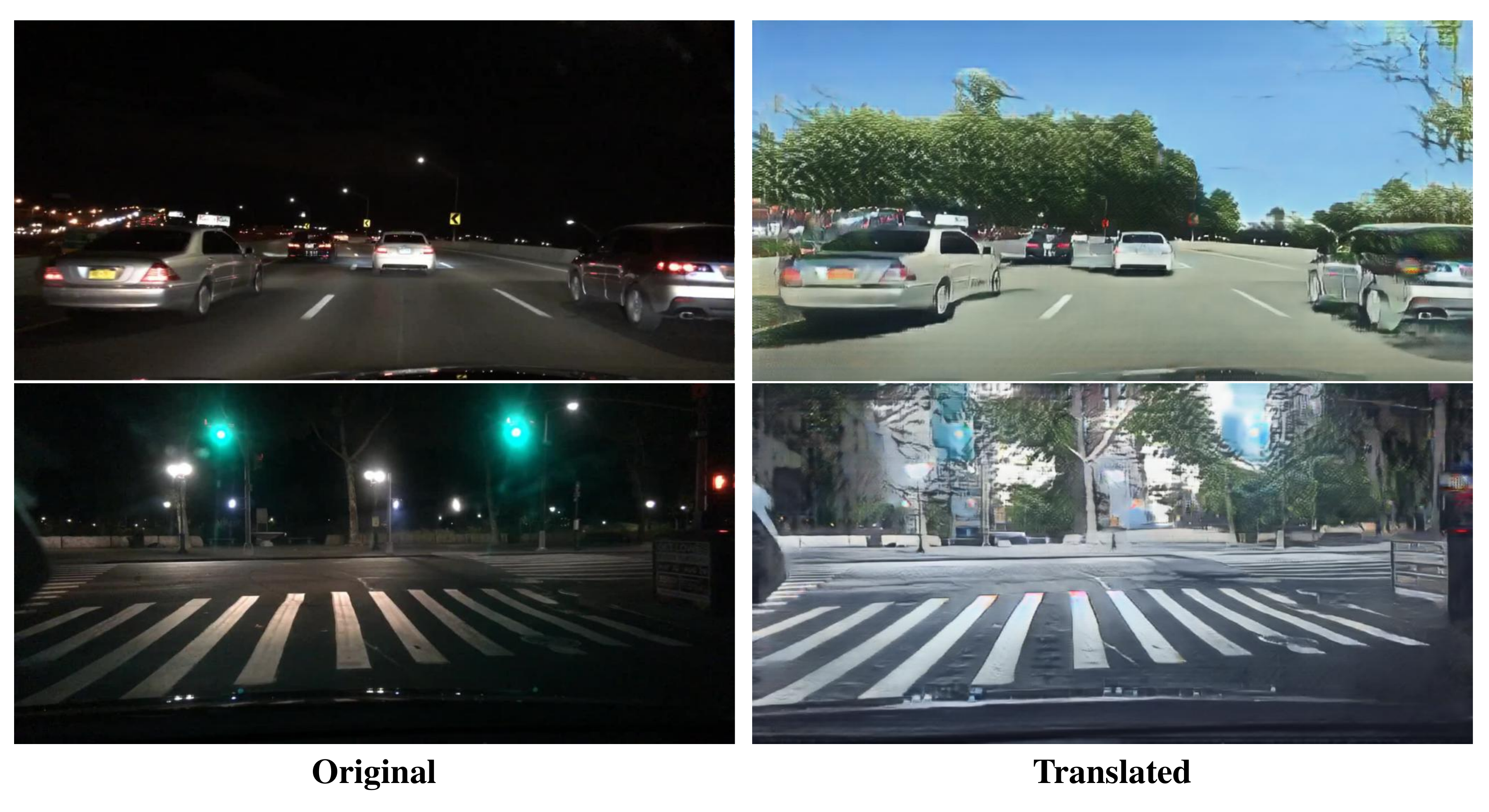}
\caption{Experiments with higher resolution ($512\times1024$) images of BDD100K} 
\label{fig:sf3}
\end{figure}  

\subsubsection{Experiments on higher resolution}
Although the resolution of train and test images in our method is $256\times512$, we additionally train our model with higher resolution images ($512\times1024$). We use BDD100K dataset only because its original resolution is $720\times1280$ (Alderley: $260\times640$). We just added one more layer in each encoder, decoder and discriminator while keeping others (\eg{} hyper-parameter and network architecture) unchanged. Although our model can translate adverse domain but it is not converged well so shows inferior visual quality and generates some artifact as shown in Fig.~\ref{fig:sf3}. We remain it for future work that finding proper hyper-parameters and network architecture to train high resolution images.

\begin{figure}[!t]
\centering \includegraphics[width=\linewidth]{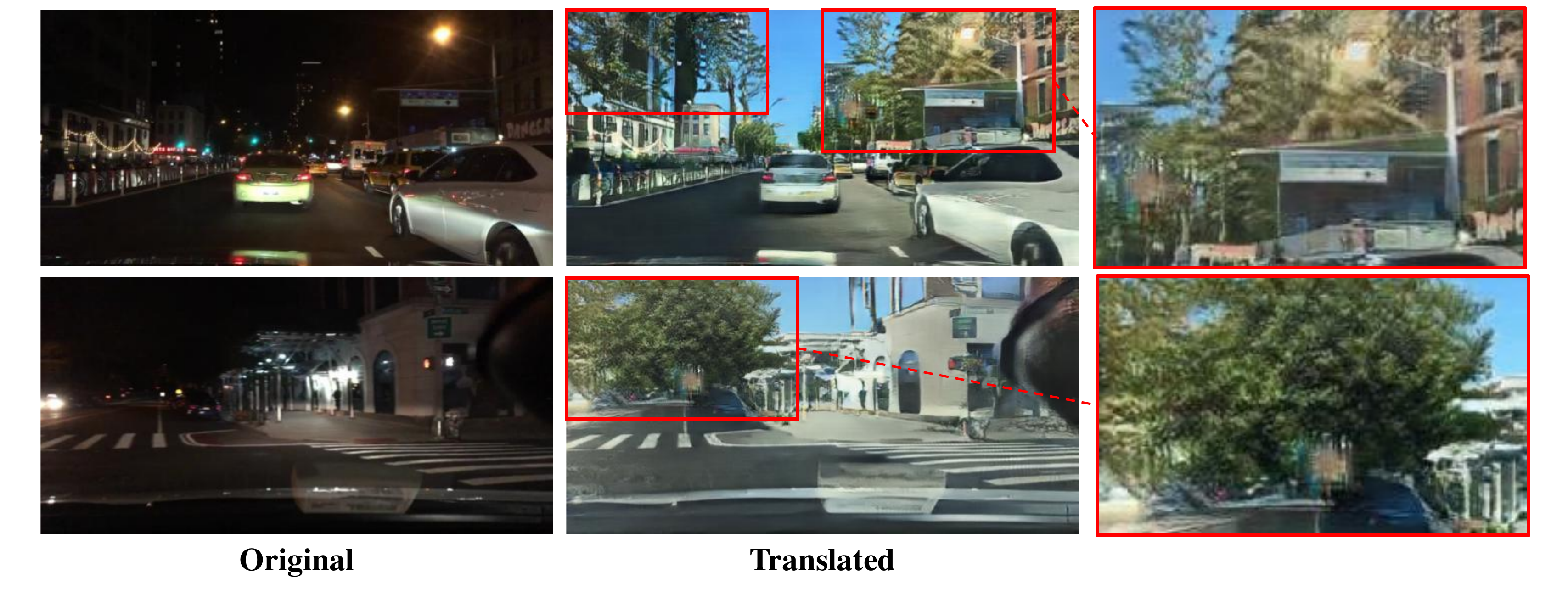}
\caption{Failure case of our method} 
\label{fig:sf4}
\end{figure}  

\subsubsection{Failure case}
We also analyze the failure case and limitation of our model. In Fig.~\ref{fig:sf4}, we show two examples of translation results (night~$\rightarrow$~day) by our model. The regions of road or car that usually appear in dataset show satisfactory translation result. However, in the case of dark building or completly dark areas, our model sometimes generates artifacts and unrealistic results such as \enquote{wooded building} or \enquote{tree on the road} (red boxes in translated results of Fig.~\ref{fig:sf4}). This is because our model is biased by dataset in that many images contain street trees. we believe that further work jointly exploiting region-based spatial attention methods with our model alleviates this problem.

\subsection{Additional comparison result and training details}
\label{sec:sec6.3}
\begin{figure}[!t]
\centering \includegraphics[width=\linewidth]{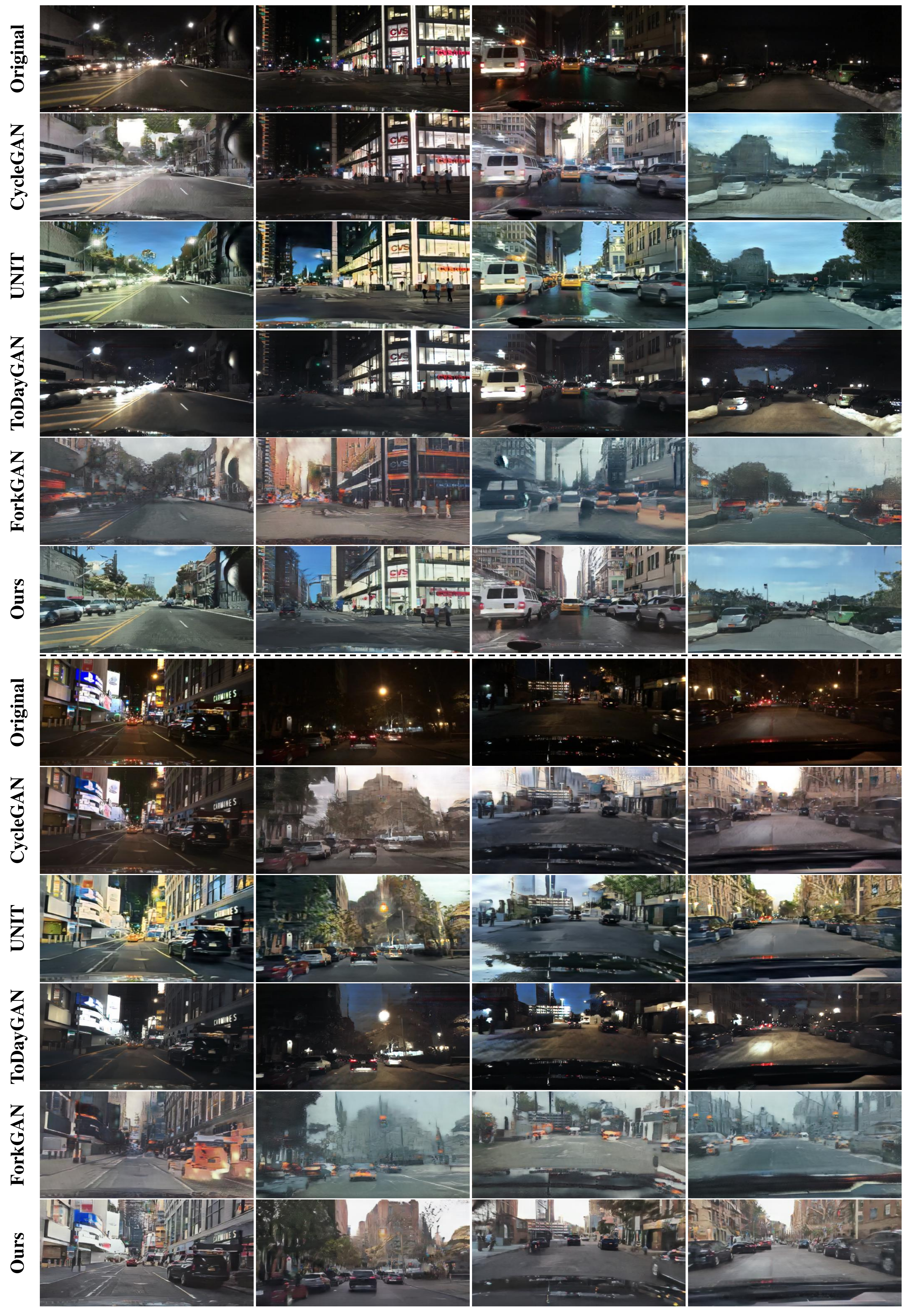}
\caption{Additional results of qualitative comparison. \textbf{Please zoom in to see more details.}} 
\label{fig:sf5}
\end{figure}

In this section, we present additional comparison of qualitative results with same methods used in the main paper and we use official implementation and settings provided by the authors, i.e., CycleGAN~\footnote{ https://github.com/junyanz/pytorch-CycleGAN-and-pix2pix}~\cite{Zhu_2017},
UNIT~\footnote{https://github.com/mingyuliutw/UNIT}~\cite{liu2017unsupervised}, ToDayGAN~\footnote{https://github.com/AAnoosheh/ToDayGAN}~\cite{anoosheh2019night}
and ForkGAN~\footnote{https://github.com/zhengziqiang/ForkGAN}~\cite{zheng2020forkgan}. All methods are trained on NVIDIA RTX Titan GPU with same datasets, i.e., Alderley~\cite{6224623} and BDD100K~\cite{yu2018bdd100k} that are cropped and resized to $256\times512$. The number of iteration for training is about 100,000 with batch size 4 and if a model could not converge and fell into mode collapse, we picked earlier checkpoint which generates reasonable results.         
As shown in Fig.~\ref{fig:sf5}, our model performs domain translation with superior visual quality while preserving objects compared to other methods. 

\subsection{Extra qualitative results}
\label{sec:sec6.4}
\begin{figure}[!t]
\centering \includegraphics[width=\linewidth]{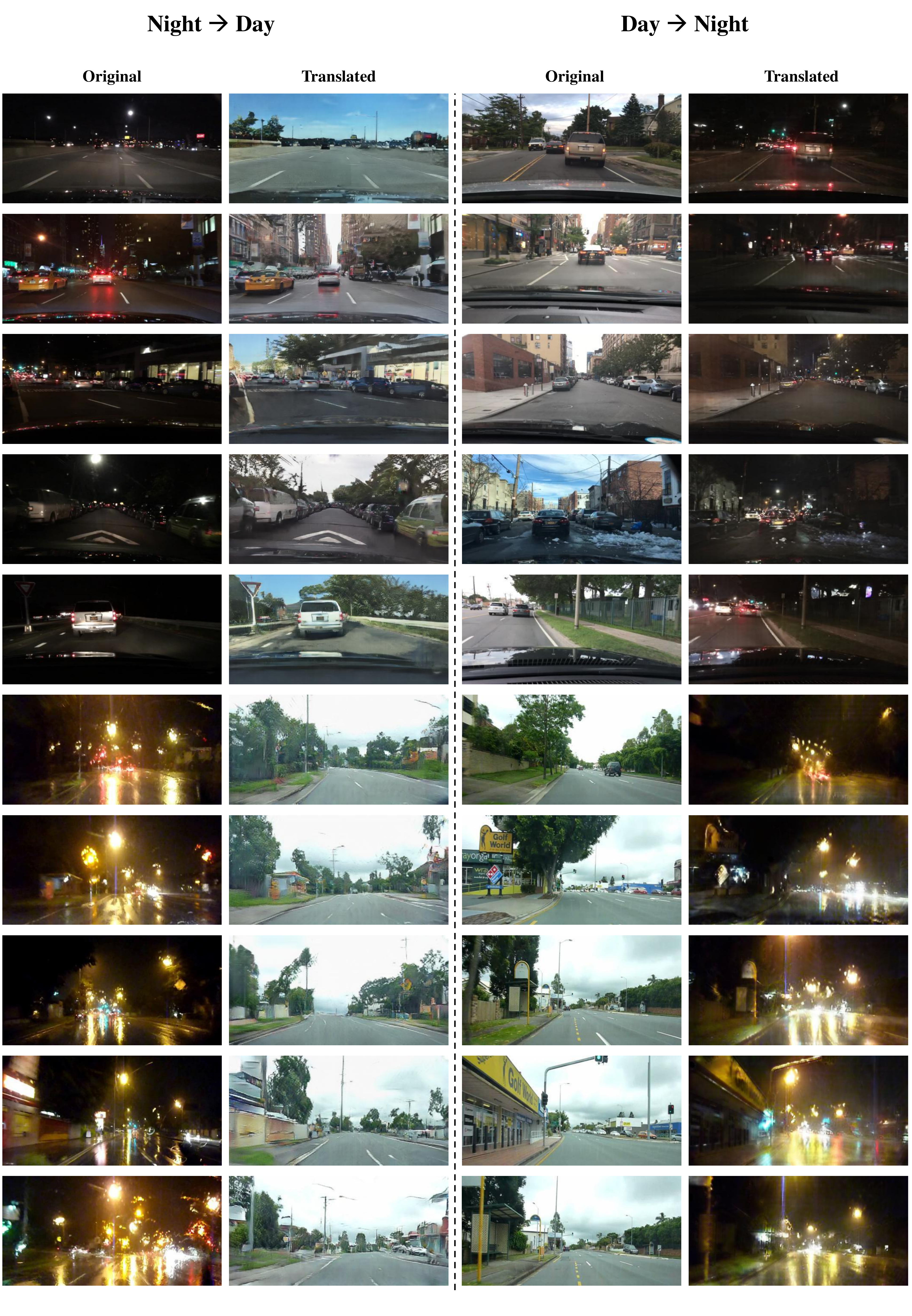}
\caption{Extra qualitative results of our model. \textbf{Please zoom in to see more details.}} 
\label{fig:sf6}
\end{figure}

Finally, we supplement the extra qualitative results (day $\leftrightarrow$ night) of our model on the datasets BDD100K~\cite{yu2018bdd100k} and Alderley~\cite{6224623}. Although the main purpose of our method is about adverse weather image translation, our model also can conduct the translation on opposite direction reasonably as shown in the right half of Fig.~\ref{fig:sf6}.  

\clearpage
\bibliography{egbib}
\end{document}